\documentclass[sigplan,screen]{acmart}

\usepackage{tikz}
\usepackage{balance}
\usepackage{amsmath}
\usepackage{verbatim}
\usepackage{xcolor}
\usepackage{amsthm}
\usepackage{amsfonts}
\usepackage{graphicx}
\usepackage{booktabs}
\usepackage{xspace}
\usepackage{hyperref}
\usepackage{url}
\usepackage[normalem]{ulem} %

\newcommand*\circled[1]{\tikz[baseline=(char.base)]{\node[shape=circle,fill,inner sep=0.5pt] (char) {\small{\textcolor{white}{#1}}};}}

\definecolor{inst}{RGB}{204, 78, 0}
\definecolor{layout}{RGB}{16, 113, 229}
\definecolor{constant}{RGB}{232, 19, 19}

\newcommand{\inst}[1]{\code{#1}}
\newcommand{\lino}[1]{({\textbf{\small Line #1}})}
\newcommand{\code}[1]{{\texttt{\small{#1}}}}
\newcommand{\dtype}[1]{\code{#1}}
\newcommand{\modsymb}{\ \%\ }
\newcommand{\system}{Tilus\xspace}
\newcommand{\vm}{\system}
\newcommand{\TritonSpeedup}{$1.75\times$\xspace}
\newcommand{\LadderSpeedup}{$2.61\times$\xspace}
\newcommand{\QuantLLMSpeedup}{$1.29\times$\xspace}
\newcommand{\MarlinSpeedup}{$1.03\times$\xspace}

\theoremstyle{definition}

\theoremstyle{remark}



\newcommand{\withUofT}[0]{
    \affiliation{ 
        \institution{University of Toronto} 
        \city{Toronto} 
        \state{ON} 
        \country{Canada}
    }
}
\newcommand{\withVector}{
    \additionalaffiliation{
      \institution{Vector Institute}
      \city{Toronto}
      \state{ON}
      \country{Canada}
    }
}
\newcommand{\withWaterloo}{
    \affiliation{
      \institution{University of Waterloo}
      \city{Waterloo}
      \state{ON}
      \country{Canada}
    }
}
\newcommand{\withAWS}{
    \affiliation{
      \institution{Amazon Web Services}
      \city{Santa Clara}
      \state{CA}
      \country{USA}
    }
}
\newcommand{\withCMU}{
    \affiliation{
      \institution{Carnegie Mellon University}
      \city{Pittsburgh}
      \state{PA}
      \country{USA}
    }
}
\newcommand{\withIndependent}{
    \affiliation{
      \institution{Independent Researcher}
      \city{Santa Clara}
      \state{CA}
      \country{USA}
    }
}
\newcommand{\withNVIDIA}{
    \additionalaffiliation{
      \institution{NVIDIA}
      \city{Santa Clara}
      \state{CA}
      \country{USA}
    }
}
\newcommand{\markNVIDIA}{\authornotemark[1]}
\newcommand{\markVector}{\authornotemark[2]}

\author{Yaoyao Ding}
\orcid{0000-0002-9801-6080}
\withUofT
\withNVIDIA
\withVector
\email{yaoyao@cs.toronto.edu}

\author{Bohan Hou}
\orcid{0000-0001-5718-3387}
\withCMU
\email{bohanhou@andrew.cmu.edu}

\author{Xiao Zhang}
\orcid{0009-0000-0914-5669}
\withUofT
\markNVIDIA
\markVector
\email{bohanhou@andrew.cmu.edu}

\author{Allan Lin}
\orcid{0009-0008-0401-1053}
\withWaterloo
\email{a8lin@uwaterloo.ca}

\author{Tianqi Chen}
\orcid{0000-0002-5744-3940}
\withCMU
\markNVIDIA
\email{tqchen@cmu.edu}

\author{Cody Hao Yu}
\orcid{0000-0002-9298-6254}
\withIndependent
\email{hao.yu.cody@gmail.com}

\author{Yida Wang}
\orcid{0000-0001-8165-840X}
\withAWS
\email{wangyida@amazon.com}

\author{Gennady Pekhimenko}
\orcid{0000-0002-3839-0919}
\withUofT
\markNVIDIA
\markVector
\email{pekhimenko@cs.toronto.edu}

\copyrightyear{2026}
\acmYear{2026}
\setcopyright{rightsretained}
\acmConference[xxx]{xxx}{xxx}{xxx}
\acmBooktitle{xxx}
\acmPrice{}

\begin{document}
\title{
Tilus: A Tile-Level GPGPU Programming Language for Low-Precision Computation
}

\begin{abstract}
Serving Large Language Models (LLMs) is critical for AI-powered applications, yet it demands substantial computational resources, particularly in memory bandwidth and computational throughput. Low-precision computation has emerged as a key technique to improve efficiency while reducing resource consumption.
Existing approaches for generating low-precision kernels are limited to weight bit widths that are powers of two and suffer from suboptimal performance because of high-level GPU programming abstractions. These abstractions restrict critical optimizations, such as fine-grained register management and optimized memory access patterns, that are essential for efficient low-precision computations.
In this paper, we introduce \textbf{Tilus}, a domain-specific language designed for General-Purpose GPU (GPGPU) computing that supports low-precision data types with arbitrary bit widths from 1 to 8 while maintaining GPU programmability. 
Tilus features a thread-block-level programming model, a hierarchical memory space, a novel algebraic layout system, and extensive support for diverse low-precision data types. 
Tilus programs are compiled into highly efficient GPU programs through automatic vectorization and instruction selection.
Extensive experiments demonstrate that Tilus efficiently supports a full spectrum of low-precision data types, and outperforms state-of-the-art low-precision kernels. 
Compared to existing compilers such as Triton and Ladder, as well as hand-optimized kernels such as QuantLLM and Marlin, Tilus achieves performance improvements of: \TritonSpeedup, \LadderSpeedup, \QuantLLMSpeedup and \MarlinSpeedup, respectively.
We open-source Tilus at \url{https://github.com/NVIDIA/tilus}.
\end{abstract}

\maketitle
\renewcommand{\shortauthors}{Yaoyao Ding et al.}

\section{Introduction}

The development of Large Language Models (LLMs) has revolutionized natural language processing tasks, enabling advanced capabilities in areas such as text generation~\cite{gpt3}, summarization~\cite{text-summarization}, translation~\cite{bloom}, and conversational AI~\cite{chatgpt}. 
However, serving LLMs poses substantial computational challenges due to the large model sizes and high computational demands. 
Efficient LLM serving demands innovative computational strategies to manage latency and power consumption constraints.
As such, optimizing LLM inference has become a priority in both industry and research to reduce latency and increase throughput of LLM serving.

Quantization~\cite{gptq, awq, quarot, quip, quip2, spin-quant} has emerged as a leading method for enhancing the efficiency of LLM serving. 
By reducing the bit-width of model parameters and activations, quantization reduces weight storage, DRAM bandwidth usage, and achieves faster computation.
For instance, A16W4 quantization (16-bit activation and 4-bit weight) reduces DRAM consumption and throughput by $4\times$ compared to the A16W16 scheme, thereby reducing the time to generate a token by about $4\times$~\cite{marlin}. 
However, the state-of-the-art 4-bit quantization methods~\cite{quarot, quip, spin-quant} still suffer from non-negligible accuracy degradation.
While using 5- to 7-bit quantization can mitigate this accuracy loss~\cite{exmy, quant-llm}, the lack of efficient GPU kernels for these bit widths has hindered their adoption.
Generating optimized kernels for hardware-unfriendly bit widths (e.g., 3, 5, 6, and 7) remains an open problem.

Existing methods for generating computation kernels fall into two main categories: manually written kernels~\cite{marlin, quant-llm} and compiler-generated kernels~\cite{tvm, ladder, triton}. 
While manually written kernels are highly optimized for specific hardware, they are time-consuming and error-prone to develop, and difficult to generalize to new architectures and evolving quantization methods. 
For example, QuantLLM~\cite{quant-llm} only supports floating-point quantization for 5- and 6-bit data types but lacks support for sub-channel quantization granularity. 
Similarly, Marlin~\cite{marlin} is limited to 4-bit signed integer quantization and does not support Hopper GPUs~\cite{hopper}.

To address these limitations, compiler-based approaches \cite{tvm, ladder, triton} have been proposed to automate kernel generation. 
Among them, Triton~\cite{triton} simplifies the GPGPU programming through a tile-based model. A Triton program defines the computation of a \emph{thread block} for tensor \emph{tiles}. 
However, Triton lacks built-in support for low-precision data types, and thus requires users to implement low-level bitwise operations manually.
Additionally, it does not expose the full GPU memory hierarchy, limiting optimization opportunities for low-precision LLM inference.
Ladder~\cite{ladder}, on the other hand, extends TVM’s scheduling system~\cite{tvm} to support low-precision computation but is restricted to data types with bit widths that are powers of two.
Moreover, its primitives cannot express crucial optimizations such as software pipelining~\cite{alcop}, leading to suboptimal performance, particularly for batch sizes greater than one during LLM decoding with \emph{continuous batching}~\cite{orca} enabled.

In response to these challenges, we propose \textbf{\system}, a tile-level GPGPU programming language backed by a virtual machine with dedicated support for low-precision computation.
\system abstracts GPU program execution into thread-block-level instructions, simplifying GPGPU programming while exposing hierarchical memory spaces for fine-grained manipulation of sub-tensors in on-chip memory. This dual approach enables efficient handling of arbitrary-precision data types while reducing the complexity of GPU programming. To achieve these goals, \system introduces:
(1) \textbf{an algebraic layout system} that specifies how tensor elements within a tile are distributed across GPU threads. 
This layout system provides a unified way to represent the storage of tile elements among different threads, making it possible to expose the registers to the kernel developers and simplify the code generation during compilation.
(2) \textbf{a thread-block-level programming model with fine-grained memory management}, providing explicit control over data movement, placement, and computation across different levels of the GPU memory hierarchy;
and comprehensive support for (3) \textbf{arbitrary low-precision data types} including signed integers, unsigned integers, and floating-point numbers with bit widths ranging from 1 to 8.

Extensive experiments show that \system extends the spectrum of efficient low-precision kernels to support arbitrary bit widths (from 1 to 8) and data type kinds (e.g., integer and floating-point numbers). At the same time, \system outperforms the state-of-the-art compilers, such as Triton~\cite{triton} and Ladder~\cite{ladder}, as well as hand-crafted kernels from QuantLLM~\cite{quant-llm} and Marlin~\cite{marlin}, achieving performance improvements of: \TritonSpeedup, \LadderSpeedup, \QuantLLMSpeedup and \MarlinSpeedup, respectively.

We summarize our key contributions as follows:
\begin{itemize}
    \item We propose a GPGPU programming language Tilus with dedicated support for low-precision computation, addressing the critical bit-width coverage and performance gap in existing approaches.
    \item Within Tilus, we introduce a novel layout system, a thread-block-level programming model with hierarchical memory space, and support low-precision data types with arbitrary bit-width from 1 to 8.
    \item Through extensive evaluation, we demonstrate that \system generates a full spectrum of highly efficient low-precision kernels, achieving up to $2.6\times$ performance improvement over state-of-the-art approaches
    on their supported kernels.
\end{itemize}

\section{Background}

\subsection{LLM Serving and Quantization}

LLM serving consists of two inference stages: prefill and decode.
The prefill stage processes the input prompt to establish context, while the decode stage iteratively generates output tokens based on prior tokens. 
Key LLM layers include multi-head attention, feed-forward networks, and layer normalization~\cite{attention}. 
Among these, matrix multiplications dominate computation time and memory consumption, making their optimization crucial for efficient LLM serving.
Quantization~\cite{llm_int8, gptq} improves their efficiency by reducing model weights and activations to lower-precision formats, such as 8-bit or 4-bit integers. 
It reduces memory usage, bandwidth requirements, and inference latency while aiming to preserve model accuracy. 
While 4-bit quantization provides significant computational savings, state-of-the-art methods~\cite{quarot, quip, spin-quant} still suffer from accuracy degradation. 
Increasing precision to 5-bit, 6-bit, or 7-bit quantization~\cite{exmy, quant-llm} can help preserve accuracy while maintaining efficiency, but these bit widths lack optimized GPU support, limiting their adoption.
Current GPU architectures and software stacks primarily optimize for power-of-two bit widths (e.g., 4-bit and 8-bit), making arbitrary bit widths computationally inefficient. 
However, demand for flexible quantization is growing, as 4-bit can be too aggressive for some models while 8-bit wastes resources. 
Supporting a broader spectrum of bit widths enables better accuracy-efficiency trade-offs in LLM serving, driving the need for new kernel generation techniques that can efficiently handle non-standard low-precision formats (e.g., those with 3, 5, 6, 7 bit widths) on modern GPUs.

\subsection{GPGPU Programming}

General-Purpose GPU (GPGPU) programming enables parallel computation by organizing tasks within a structured execution and memory hierarchy~\cite{gpu_era}. 
The execution hierarchy begins with the \emph{thread}, the smallest unit of execution, which performs instructions independently, using its own registers and local memory. 
Threads are grouped into \emph{thread blocks}, which enable data sharing through shared memory and support synchronized execution.
A \emph{grid} consists of multiple independent thread blocks, enabling large-scale parallelism by organizing thousands or millions of threads.
The GPU memory hierarchy comprises registers, shared memory, and global memory.
\emph{Registers} provide the fastest and thread-private storage. 
\emph{Shared memory} is accessible by all threads within a thread block and faster than global memory. 
\emph{Global memory} is accessible across the entire grid with high latency. 
This structure allows for highly efficient parallel execution by leveraging both the execution and memory hierarchies.

\subsection{The GPGPU Languages and Compilers}

\subsubsection{GPGPU Programming Languages}

GPGPU programming involves various languages and compilers that balance hardware abstraction with control. 
Low-level languages like SASS~\cite{sass}
and CDNA3~\cite{cdna3} offer direct hardware access for fine-grained optimizations but require deep architectural knowledge. 
Slightly higher in abstraction, NVIDIA's PTX~\cite{ptx} serves as an intermediate representation that links high-level languages like CUDA~\cite{cuda} to GPU-specific instructions while preserving optimization flexibility. 
High-level languages like CUDA~\cite{cuda} and HIP~\cite{hip} simplify programming by extending the C programming language. 
Despite these languages, GPGPU programming remains complex. It is constrained by hardware-specific memory and computation hierarchies and requires workload-specific optimizations. 
To address these challenges, researchers have introduced higher-level languages and compilers, classified into two categories: \emph{tile-oriented compilers}, which simplify programming through abstractions beyond CUDA~\cite{cuda}, and \emph{schedule-oriented compilers}, which optimize computation-hardware mappings via declarative scheduling primitives.

\subsubsection{Tile-Oriented Compilers}

This type of compilers, such as Graphene~\cite{graphene}, Hidet~\cite{hidet}, and Triton~\cite{triton}, 
enables programmers to write kernels directly, offering abstractions like tile types or tile-level task/element distribution to simplify the process. Triton~\cite{triton}, for instance, introduces the tile programming model, where thread block behavior is defined programmatically, and tiles replace scalars as the basic data type. This approach combines programming simplicity with high-performance kernel generation, making Triton widely adopted.
However, Triton lacks native support for low precision data types like \dtype{uint4}. Handling these types requires manually unpacking sub-byte data from larger storage types (e.g., \dtype{uint32})~\cite{triton_a16w4}. Additionally, Triton does not expose the GPU memory hierarchy, limiting programmers' control over data loading and memory scope usage, which complicates performance optimization for low-precision kernels. These limitations result in inefficient low-precision kernel execution.
Figure~\ref{fig:weight-pipeline}(a) illustrates the inefficiencies in Triton-generated low-precision kernels, using a \dtype{uint4} weight loading pipeline as an example. The process includes four steps: \circled{1} weights are asynchronously copied from global memory to shared memory using pipelined \code{cp.async} instructions~\cite{alcop}; \circled{2} shared memory data is loaded into registers; \circled{3} unpacking and casting operations are performed; and \circled{4} the register tensor layout is converted to meet the requirements of tensor core instructions. Among these, step \circled{4} is a major bottleneck due to the reliance on shared memory for layout conversion, which incurs significant overhead.

\subsubsection{Schedule-Oriented Compilers}

\begin{figure}[t]
    \centering
    \includegraphics[width=1.0\linewidth]{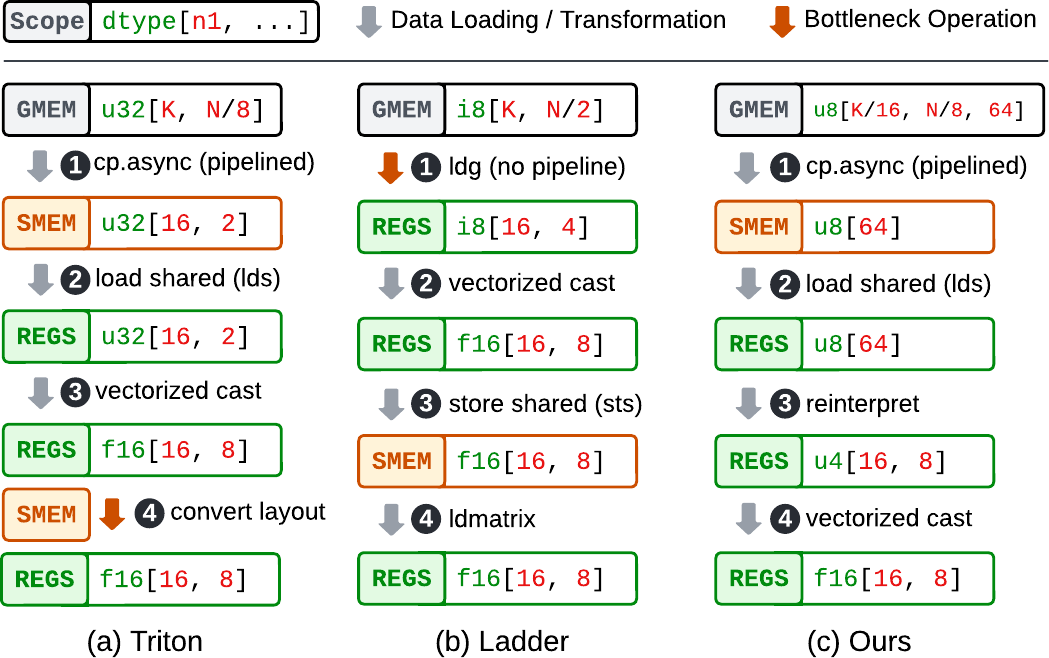} 
    \Description{Weight loading pipeline illustration}
    \caption{The weight loading pipeline of Triton, Ladder, and our approach. The tensors could be in global memory (GMEM), shared memory (SMEM), or registers (REGS). }%
    \label{fig:weight-pipeline}%
\end{figure}

Schedule-oriented compilers decouple computation from scheduling to optimize the computation-to-hardware mappings. Halide~\cite{halide} pioneered this approach, which was later extended by TVM~\cite{tvm} and subsequent works~\cite{autotvm, ansor, amos, meta-schedule, alcop, tensorir, ladder} in the domain of deep learning. Among them, Ladder~\cite{ladder} is the first one to support low-precision computation by introducing dedicated primitives to pack low-precision data (e.g., 4-bit integers) into larger types (e.g., 8-bit integers).
However, Ladder~\cite{ladder} has two limitations. First, it cannot handle non-power-of-two bit widths efficiently due to \emph{type-level packing}, packing low-precision types into storage types. Second, its primitive-style scheduling prevents optimizations like \emph{software pipelining}~\cite{alcop}, resulting in suboptimal performance. Figure~\ref{fig:weight-pipeline} (b) illustrates the weight loading process in Ladder's low-precision kernels. This process includes \circled{1} loading weights from global memory to registers without pipelining; \circled{2} vectorized casting; \circled{3} storing the cast results in shared memory; and finally \circled{4} using the \code{ldmatrix} instruction to load weights from shared memory to registers for subsequent tensor core operations. This lack of pipelining between weight loading and computation significantly hinders performance.

\begin{figure*}[ht!]%
    \centering
    \includegraphics[width=1.0\linewidth]{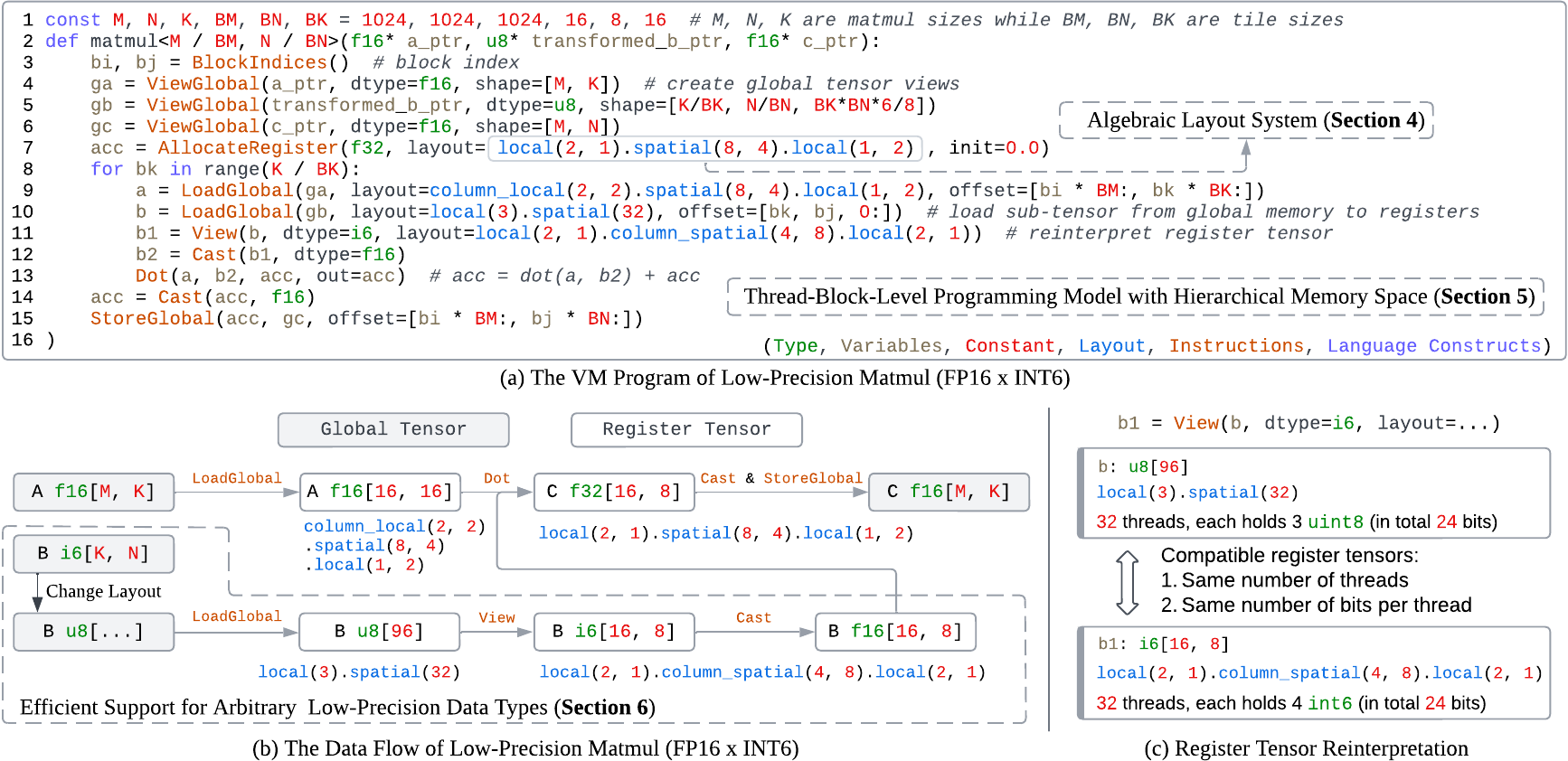} 
    \caption{
    This figure provides a concrete example of how the Tilus is used to implement low-precision matrix multiplication (\code{FP16} $\times$ \code{INT6}).
    Figure (a) illustrates the virtual machine program, highlighting key features such as the algebraic layout system (Section~\ref{section:system:layout}), thread-block-level instructions (Section~\ref{section:system:instruction-set}), and efficient low-precision data support.
    Figure (b) illustrates the kernel’s data flow, emphasizing tensor movement across the memory hierarchy and intermediate operations such as tensor reinterpretation and type casting.
    This similar weight-loading strategy can be applied to arbitrary type widths (Section~\ref{section:system:low-precision}).
    Finally, Figure (c) demonstrates register tensor reinterpretation, showing how tensors with compatible bit distributions across threads (e.g., 24 bits per thread) can be efficiently reinterpreted into different data types and layouts.
    }%
    \Description{matmul example.}
    \label{fig:overview}%
\end{figure*}

\section{System Overview}

\subsection{Key Ideas}

Our work introduces a domain-specific language, Tilus, that provides fine-grained control over shared memory and registers, making it possible to program efficient low-precision deep learning kernels. 
Tilus supports low-precision data types with arbitrary bit widths ranging from 1 to 8, enabling efficient weight loading and computation.
Figure~\ref{fig:weight-pipeline} (c) shows the weight loading pipeline of a Tilus program, using \dtype{uint4} as an example. 
It begins with \circled{1} a pipelined asynchronous memory copy from global memory to shared memory, followed by \circled{2} loading the register tensor from shared memory. Next, it \circled{3} \emph{reinterprets} the register tensor into a different data type and layout at no cost, before finally \circled{4} performing vectorized casting. 
This pipeline achieves superior efficiency compared to the other methods in Figure~\ref{fig:weight-pipeline}, as it eliminates layout conversion (unlike Triton~\cite{triton}) and incorporates pipelining (unlike Ladder~\cite{ladder}).
More importantly, our pipeline is generic, making our work the first to seamlessly support arbitrary low-precision data types with bit widths ranging from 1 to 8 bits.

To achieve this efficiency, our design is built on several key ideas.
\textbf{A GPGPU Virtual Machine}:
every Tilus program is a program for an abstract GPGPU virtual machine (VM) that contains an instruction set.
This decision stems from the need for greater flexibility in GPU programming. By abstracting GPU functionalities, such as memory loading and computation, into instructions, it becomes easier to add support for new architectural features while keeping support for older ones.
\textbf{A Thread-Block-Level Programming Model with Hierarchical Memory Spaces}:
The underlying VM explicitly exposes the GPU memory hierarchy --- including registers, shared memory, and global memory --- that existing solutions like Triton~\cite{triton} abstract away. By granting programmers fine-grained control over data placement and movement, our approach enables memory pipelining and eliminates unnecessary layout conversions, as shown in Figure~\ref{fig:weight-pipeline}.
\textbf{An Algebraic Layout System}:
we introduce an algebraic layout system that precisely defines how elements within a register tensor are distributed among threads. This structured representation simplifies the construction, analysis, and interpretation of tensor layouts. Notably, it enables seamless reinterpretation of low-precision register tensors into standard data types, as demonstrated in Step \circled{3} of Figure~\ref{fig:weight-pipeline}(c).
\textbf{Native Support for Arbitrary Low-Precision Data Types}:
Tilus provides built-in support for a wide range of low-precision data types, including both signed and unsigned integers and floating-point numbers with bit widths from 1 to 8. Supported types include \dtype{int2} to \dtype{int8}, \dtype{uint1} to \dtype{uint8}, and \dtype{float3} to \dtype{float8}, with arbitrary \emph{exponent} and \emph{mantissa} distribution for floating-point types. 
These innovations collectively enhance the programmability, efficiency, and flexibility of low-precision kernel development on modern GPUs.
We chose not to extend Triton~\cite{triton} because its programming model inherently abstracts away tensor layouts, making it incompatible with our approach of explicit layout control. Similarly, Ladder~\cite{ladder} relies on type-level packing, whereas \system employs tile-level reinterpretation, making the two fundamentally incompatible.
The next section presents a Tilus-programmed example of low-precision matrix multiplication.

\subsection{An Example of Tilus Program}
Figure~\ref{fig:overview} illustrates a low-precision matrix multiplication in Tilus.
Matrix multiplication is defined as $C_{M, N} = A_{M,K} \times B_{K,N}$, where $A$ and $B$ are \dtype{float16} (a 16-bit floating-point number~\cite{ieee_float}) and \dtype{int6} (a 6-bit signed integer), respectively.
The kernel performs matrix multiplication with dimensions \code{M}, \code{N}, and \code{K}, where each thread block computes a \code{BM} $\times$ \code{BN} tile of the C matrix~\lino{1}.
Therefore, a grid of (\code{M / BM}, \code{N / BN}) thread blocks must be launched~\lino{2}.
Inside the kernel, the \inst{BlockIndices} instruction retrieves the thread block indices \code{bi} and \code{bj}~\lino{3}, which determine the offset (\code{bi * BM}, \code{bj * BN}) for computing the corresponding C tile.
Three tensor views are created for the input and output tensors in global memory by specifying their addresses and shapes~\lino{4-6}.
Then, a register tensor of type \code{f16[16, 8]} is created with the following layout:
\begin{center}
\code{\textcolor{layout}{local}(\textcolor{constant}{2}, \textcolor{constant}{1}).\textcolor{layout}{spatial}(\textcolor{constant}{8}, \textcolor{constant}{4}).\textcolor{layout}{local}(\textcolor{constant}{1}, \textcolor{constant}{2})}.
\end{center}
It distributes $16 \times 8 = 128$ elements across 32 threads. Each thread stores 4 elements~\lino{7}.
This layout is \emph{composed} of three \emph{primitive layouts} (Section~\ref{section:system:layout}) and aligns with the C matrix layout used by the \code{mma.m16n8k16} tensor core instruction in PTX~\cite{ptx}.
The reduction loop over the k dimension~\lino{8-13} repeatedly loads tiles of A and B from global memory into registers and accumulates their product.
For each iteration, we first load a \code{f16[16, 16]} tile from global memory to register with a \inst{LoadGlobal} instruction~\lino{9}. 
The layout of the loaded register tile is specified and required by the tensor core instruction.
The offset parameter specifies the position of the loaded tile within the global tensor.
Loading tensor B, with data type \dtype{int6}, involves a more complex process, detailed in Section~\ref{section:system:low-precision}.
We summarize the high-level ideas here.
As a pre-processing step before launching the kernel, the weight tensor's layout in global memory is transformed from \code{i6[K, N]} to \code{u8[K / BK, N / BN, BK * BN * 6 / 8]}, enabling efficient loading via the \inst{LoadGlobal} instruction (the `Change Layout' step in Figure~\ref{fig:overview} (b)).
Next, in the kernel, the transformed tile is loaded into a register tensor~\lino{10} and then \emph{reinterpreted} to a tensor with a different data type and layout~\lino{11}.
This reinterpretation is valid because both tensors are stored across the same number of threads (32), with each thread holding exactly 24 bits: 3 $\times$ \code{u8} or 4 $\times$ \code{i6}, as shown in Figure~\ref{fig:overview} (c).
Following this, the \code{i6} tensor is cast to an \code{f16} tensor~\lino{12}, which is then fed to the tensor core to perform matrix-multiply accumulate (mma)~\lino{13}.
Finally, the accumulation tensor is cast from \code{f32} to \code{f16} and stored in global memory~\lino{14-15}.
For simplicity, this program does not use shared memory and omits optimizations like software pipelining~\cite{alcop}. Additionally, each k-iteration performs only a single tensor core instruction~\cite{cuda}.

The following sections introduce the three core components of Tilus.
Section~\ref{section:system:layout} introduces an algebraic layout formulation to systematically define how the elements of a tile are stored in the registers among the block threads. Section~\ref{section:system:instruction-set} introduces the thread-block-level programming model with a hierarchical memory space exposed explicitly. Section~\ref{section:system:low-precision} introduces the native support for arbitrary low-precision data types to address the growing demand for low-precision computation in deep learning workloads.

\section{Algebraic Layout System}
\label{section:system:layout}
\begin{figure}[th]%
    \centering
    \includegraphics[width=1.0\linewidth]{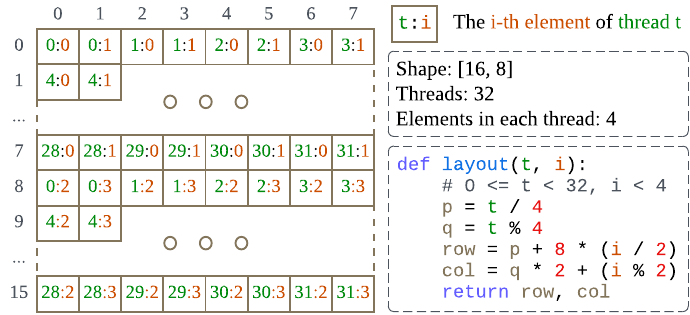} 
    \caption{
    Layout of operand A in a Tensor Core instruction. The operand, with $16 \times 8$ elements, is distributed across $32$ threads, with each thread storing four elements. The logical index of each element is determined by a layout function given the thread index \code{t} and the local element index \code{i}.
    }%
    \Description{mma layout}
    \label{fig:mma_layout}%
\end{figure}
Tilus exposes a hierarchical memory space to programmers, comprising global memory, shared memory, and registers.
We need a way to model the mapping between the logical index of a tensor element and the location of the corresponding element in memory for all three memory scopes.
Such a mapping is usually called the \emph{layout} of the tensor.
Of the memory scopes, the layout for register tensors is the most complicated.
Figure~\ref{fig:mma_layout} illustrates an example of the layout used by a tensor core instruction: \code{mma.m16n8k8.f32.f16.f16.f32 D, A, B, C}.
It performs the following computation:
$
D_{16, 8} = A_{16, 8}B_{8, 8} + C_{16, 8}
$
where $A, B, C, D$ are tensors stored in thread registers and distributed across the $32$ threads in a warp.
Since the elements are spread across different threads, we refer to this layout as a \emph{distributed layout}~\cite{triton}.
Such a layout can be defined as a function $f$ that maps a thread index $t$ and a local index $i$ to the logical index $f(t, i)$ of the tensor element. 
For example, the layout in Figure~\ref{fig:mma_layout} can be represented as:
$$
f(t, i) = (t / 4 + i / 2 \times 8, t \modsymb 4 \times 2 + i \modsymb 2)\footnotemark
$$
Here, $t$ ranges from $0$ to $31$, and $i$ from $0$ to $3$.
The function $f(t, i)$ represents the logical index of the element stored in the local element $i$ in thread $t$.

\subsection{Parameterized Primitive Layouts}
\begin{figure}[th]%
    \centering
    \includegraphics[width=0.90\linewidth]{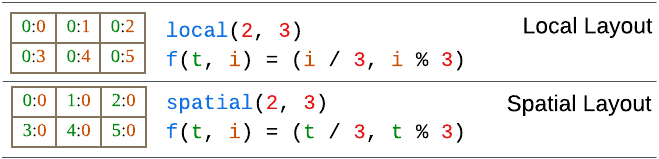} 
    \caption{
    Two types of primitive layouts: local and spatial. A local layout stores all tile elements within a single thread, whereas a spatial layout distributes them across multiple threads, with each thread holding only a single element.
    }%
    \Description{primitive layouts}
    \label{fig:primitive_layout}%
\end{figure}
With the formal definition of layout, we introduce parameterized primitive layouts that serve as the fundamental building blocks of our layout algebra. Given a tile\footnote{We use the terms \emph{tile} and \emph{tensor} interchangeably.} with shape \((n_1, n_2)\), there are two primary ways to store it: (1) to store all \(n_1n_2\) elements in a single thread, or (2) to distribute all elements across \(n_1n_2\) threads, with each thread holding a single element. We refer to the first type as \emph{local layouts}, denoted as \code{local(n1, n2)}, and the second as \emph{spatial layouts}, denoted as \code{spatial(n1, n2)}. This concept naturally extends to tiles with arbitrary dimensions.
Figure~\ref{fig:primitive_layout} illustrates these two primitive layouts. The \code{local(2, 3)} layout maps the \(i\)-th local element of thread \(t\) to the logical index \((i / 3, i \modsymb 3)\), while the \code{spatial(2, 3)} layout maps it to \((t / 3, t \modsymb 3)\).
We observe that the layouts for all common operators in LLMs can be constructed using these two primitive layouts.
In the next section, we introduce the Kronecker product of layouts to construct more complex layouts.

\subsection{Kronecker Product}
\begin{figure}[th]%
    \centering
    \includegraphics[width=1.0\linewidth]{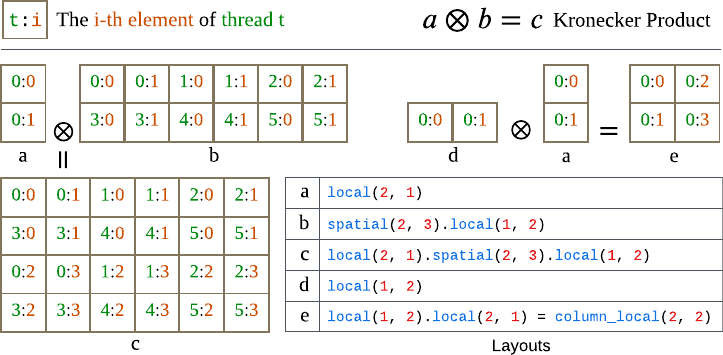} 
    \caption{
    Examples of Kronecker products over layouts. In the figure, layout (c) is the product of layouts (a) and (b), while layout (e) is the product of layouts (d) and (a).
    }%
    \Description{layout composition}
    \label{fig:layout_composition}%
\end{figure}

The layouts used in modern deep learning workloads, as well as those defined by hardware instructions, typically exhibit a hierarchical structure.
Consider layout (c) in Figure~\ref{fig:layout_composition} as an example.
This layout has shape $(4,6)$, storing $24$ elements across $6$ threads. Each thread holds four elements.
We denote the four elements stored in each thread as $a_0, a_1, a_2, a_3$.
Comparing its first two rows with the last two, we observe a similar structure, except that the last two rows store elements in $a_2$ and $a_3$ instead of $a_0$ and $a_1$.
To model this structural invariance, layout (c) can be viewed as a \emph{Kronecker product} of layouts (a) and (b), with each element in layout (a) representing a tile with layout (b).
Indeed, layouts (a) and (b) can be multiplied to represent layout (c):
\begin{align*}
c(t, i) & = a(t / 6, i / 2)\odot (2, 6) + b(t\modsymb 6, i \modsymb 2),
\end{align*}
where $0 \leq t < 6$, $0 \leq i < 4$, $\odot$ denotes the element-wise product, and $(2, 6)$ represents the shape of layout (c).
The Kronecker product can be generalized.
Given two layouts $f$ and $g$ with the same number of dimensions, we define their Kronecker product, $h=f\otimes g$ as 
\begin{align*}
h(t, i) = f(t / T_g, i / N_g)\odot S_g + g(t \modsymb T_g, i \modsymb N_g),
\end{align*}
where $T_g, N_g, S_g$ represent the number of threads, the number of local elements per thread, and the shape of layout $g$, respectively.
We can prove that this operation is \emph{associative}, meaning that for any three layouts $f$, $g$, and $h$, the equality $f\otimes(g\otimes h) = (f\otimes g)\otimes h$ holds. 
However, the operation is not commutative, meaning that, in general, $f\otimes g \neq g\otimes f$.
Spatial and local layouts follow a row-major ordering for threads and local elements, respectively. 
Using this operation, we can construct their column-major counterparts, \code{column\_spatial(...)} and \code{column\_local(...)}, as demonstrated by layout (e) in Figure~\ref{fig:layout_composition}.
Returning to the tensor core instruction layout in Figure~\ref{fig:mma_layout}, it can be expressed as a Kronecker product \code{local(2, 1).spatial(8, 4).local(1, 2)}.
We can also define its inverse operation. If $h=f\otimes g$, we define $f=h / g$ as the result of dividing layout $h$ by layout $g$. 
For example, dividing \code{local(2, 4)} by \code{local(1, 2)} results in \code{local(2, 2)}. 
We also refer to the Kronecker product as layout composition in some contexts.

\section{Unified Layout Representation}
\label{sec:unified-layout-representation}
\begin{figure}[th]%
    \centering
    \includegraphics[width=1.0\linewidth]{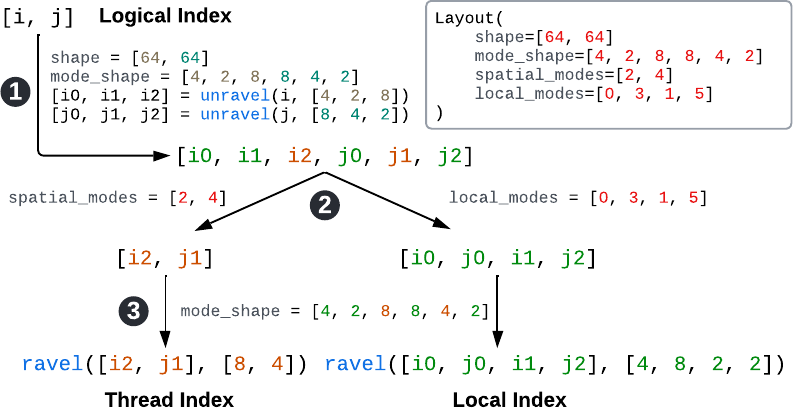} 
    \caption{Example of the unified layout representation.}%
    \Description{}
    \label{fig:layout-representation}%
\end{figure}

We use a unified representation for all layouts of register tensors in Tilus. This representation gives each layout four attributes: \code{shape}, \code{mode\_shape}, \code{spatial\_modes}, and \code{local\_modes}.
The \code{shape} is a sequence of integers that defines the shape of the register tensor.
We can split each dimension of the register tensor into some sub-dimensions (we call them mode following prior work~\cite{graphene}) and concatenate these sub-dimensions to get the \code{mode\_shape}. 
Then we use \code{spatial\_modes} and \code{local\_modes} to specify the sub-dimensions assigned to spatial threads and to the local storage of each thread.
The dimension split-distribute-merge method uniquely defines a register layout.

Figure~\ref{fig:layout-representation} shows an example of a layout and how we map the logical index of a register tensor element to the pair of \code{thread\_index} and \code{local\_index}.
There are three steps: 1) split dimensions, 2) distribute sub-dimensions; and 3) merge sub-dimensions. 
Given a logical index \code{[i, j]}, we first \circled{1} split each index into the indices of its sub-dimensions (i.e., \code{[i0, i1, i2]} for \code{i} and \code{[j0, j1, j2]} for \code{j}) with the \code{unravel} operation.
After that, we \circled{2} distribute the sub-dimension indices to get the indices for spatial threads (i.e., \code{[i2, j1]}) and local storage (\code{[i0, j0, i1, j2]}). 
Finally, we \circled{3} convert the multi-dimensional indices for threads and local storage into linear index to get the thread index and local index. 
The \code{ravel} and \code{unravel} functions are used to convert between multi-dimensional index in a grid with given shape and its row-major linear index. 
For example: 
\code{unravel(i, [4, 2, 8])} = \code{[i / 16, i / 8 \% 2, i \% 8]}, and
\code{ravel([i2, j1], [8, 4])} = \code{ i2 * 4 + j1}.

The layouts represented in this form are closed under the Kronecker product, meaning that the product result of two layouts in this form can also be represented in this form.

\section{Thread-Block-Level Programming Model}
\label{section:system:instruction-set}

Modern GPU programming models, such as PTX~\cite{ptx} and CUDA~\cite{cuda}, define operations at the thread level, following the Single-Instruction-Multiple-Thread (SIMT) paradigm~\cite{simt}.
To simplify GPU programming, we adopt the thread-block-level programming model, defining operations at the granularity of thread blocks rather than individual threads.
Additionally, building on the layout system introduced previously, we propose explicitly exposing the hierarchical memory structure in modern GPUs, enabling fine-grained memory control while reducing programming complexity.
We refer to this model as Single-Instruction-Multiple-Block (SIMB). In this section, we will introduce the type system, program structure, and instruction set of Tilus.

\subsection{State Space and Type System}
Tilus supports three variable types.
\emph{Scalar variables} store individual values, such as integers (e.g., \dtype{int32}) or floating-point numbers (e.g., \dtype{float16}).
\emph{Pointer variables} store memory addresses rather than direct data values.
\emph{Tensor variables} represent multi-dimensional arrays, with types that specify their shape, element type, memory scope, and layout.
Tensors reside in different memory scopes, including global memory, shared memory, and registers.
The tensor layout determines how high-dimensional tensor elements are mapped to linear memory.
All variables in Tilus operate at the thread-block level, meaning that all threads within a block collaboratively allocate and maintain their state.

\subsection{Program Structure and Control Flow}
\begin{figure}[th]%
    \centering
    \includegraphics[width=1.0\linewidth]{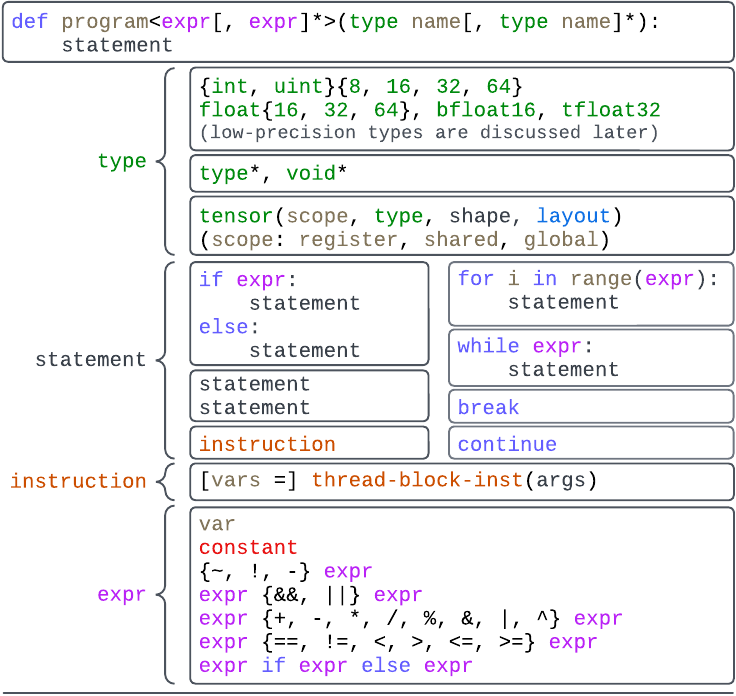} 
    \caption{A Tilus program contains parameters and a body. The body is a list of control-flow statements or block-level instructions. 
    The majority of functionality, such as tensor allocation and computation, is provided by instructions.
    }%
    \Description{instruction}
    \label{fig:program-structure}%
\end{figure}
\begin{table*}[ht]
    \centering
    \caption{The thread-block-level instruction set of Tilus's virtual machine. 
    Each instruction specifies an operation applied to the entire thread block. 
    Parameters enclosed in \code{[...]} are optional. 
    Instructions that return a new register tensor also have an in-place variant, which writes the result to an existing register tensor using the \emph{out} parameter instead of creating a new tensor.
    }%
    \includegraphics[width=1.0\linewidth]{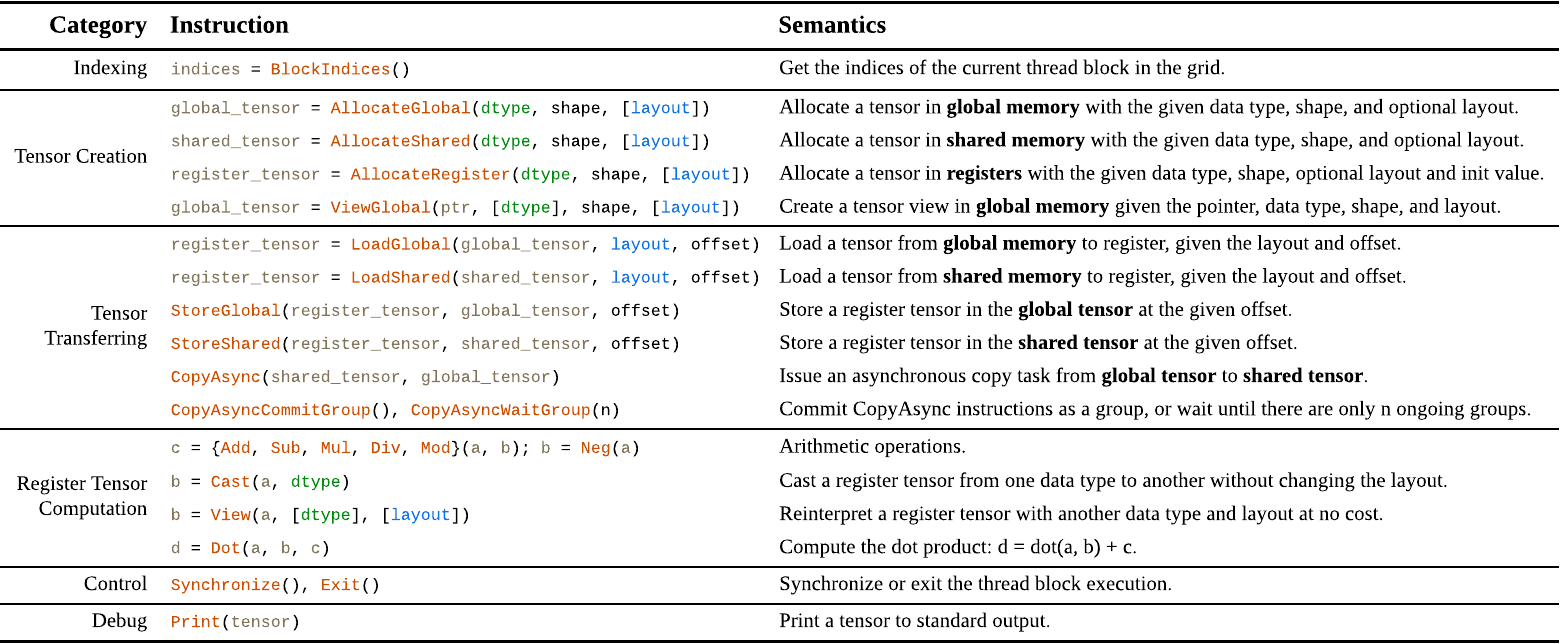} 
    \label{table:instruction-set}%
\end{table*}
Figure~\ref{fig:program-structure} illustrates the structure of a Tilus program.
Each program consists of a program name, a grid shape, a list of parameters, and a program body.
The grid shape is specified as a list of expressions enclosed in \code{<...>}, where each expression is either a positive integer or an integer expression based on the program parameters.
If the grid shape contains parameter-based expressions, its dimensions are determined at runtime based on the program’s launch arguments.
The program body consists of a sequence of statements, including if-else statements, range-based for-loops, and while-loops.
Unlike other low-level virtual machines~\cite{ptx} or instruction set architectures (ISAs)~\cite{sass}, our virtual machine does not abstract control-flow statements into jump instructions. Instead, it retains high-level control structures to improve readability and ease of programming for human developers.
In addition to control-flow statements, individual instructions can also serve as statements. 
Most of Tilus's functionality is implemented as instructions within its instruction set.

\subsection{Thread-Block-Level Instruction Set}
Each instruction in the Tilus's instruction set operates at the thread-block level rather than the thread level.
Table~\ref{table:instruction-set} shows a list of the instructions in the instruction set, with the signature of each instruction and a brief description of the instruction semantics.
These instructions allocate tensors with specific data types, shapes, and layouts in designated memory spaces (e.g., global memory, shared memory, registers), transfer tensors between memory spaces, and perform computations or transformations on register tensors.
The execution model of modern GPUs allows different warps to execute different instructions at the same time.
Similarly, the execution of instructions in Tilus exhibits this behavior: certain subsequent instructions may begin execution before the current instruction completes, resulting in multiple block-level instructions being in-flight simultaneously.
Generally, this behavior does not pose significant issues. 
However, an exception occurs when two instructions access the same region of shared or global memory, and the second instruction depends on the completion of the first. 
In such cases, a \code{Synchronize} instruction must be inserted to ensure all preceding instructions complete before subsequent ones execute. 
Instructions like \code{Print} are used for debugging.

\section{Arbitrary Low-Precision Data Types}
\label{section:system:low-precision}
Modern processors use bytes (8 bits each) as the smallest processing unit.
As a result, standard data types in modern programming languages typically have bit widths that are multiples of 8.
However, the high computational and memory demands of LLMs make low-precision data types with less than 8 bits essential for reducing resource consumption.
This section describes how Tilus efficiently supports low-precision data types with bit width from 1 to 8.

\subsection{Storage of Low-Precision Data}
\begin{figure}[th]%
    \centering
    \includegraphics[width=1.0\linewidth]{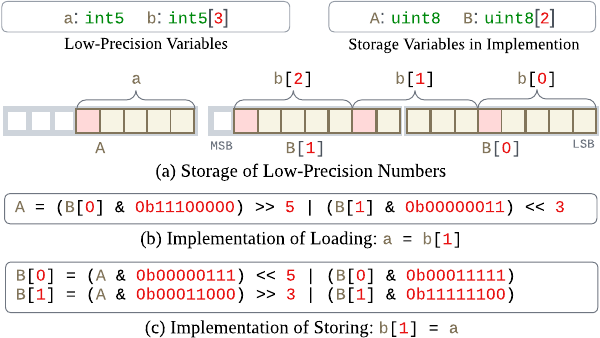} 
    \caption{
    Compact storage and access of low-precision data.
    Figure (a) illustrates the use of the \code{uint8} type to store low-precision data, where some elements may span two consecutive bytes.
    Figure (b–c) illustrate the implementation of loading and storing low-precision elements.
    }%
    \Description{}
    \label{fig:storage-and-access}%
\end{figure}
\label{section:system:low-precision-storage}
Since modern processors, including CPUs and GPUs, use bytes as the smallest unit for memory access and computation, we store low-precision data (fewer than 8 bits per element) \emph{compactly} within bytes, as shown in Figure~\ref{fig:storage-and-access}.
Compact storage eliminates bit gaps between consecutive low-precision values, which may result in a single value spanning two \code{uint8} entries (e.g., \code{b[1]} in Figure~\ref{fig:storage-and-access}).
Bitwise operations are employed to extract, manipulate, and store low-precision values within packed byte arrays.
To load a low-precision value, we first extract relevant bits using bitwise AND, adjust their position with bitwise SHIFT operations, and finally combine separated parts using bitwise OR if the value spans multiple bytes.
Similarly, to store a low-precision value, we first clear the target bit positions using a bitwise mask, then insert the new value using bitwise OR while preserving the other bits.
Low-precision data is cast to standard data types before arithmetic computations and is cast back afterward.
While these methods enable support for arbitrary bit-width data types, they are often inefficient. They serve only as a fallback mechanism.
More efficient handling of low-precision data is necessary for LLM serving.

\subsection{Efficient Low-Precision Support in LLMs}
\label{section:low-precision:efficient-support-in-llm}
\begin{figure}[th]%
    \centering
    \includegraphics[width=1.0\linewidth]{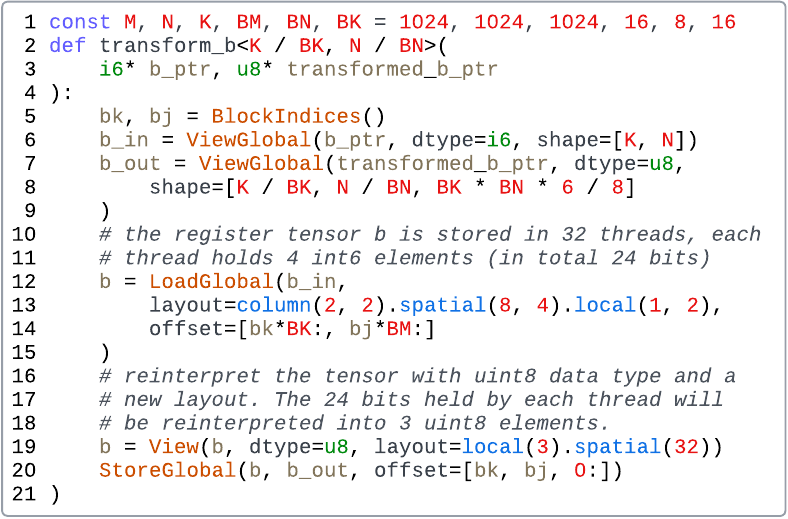} 
    \caption{
    Program to rearrange tensor B with data type \code{int6}, used in the "Change Layout" step of Figure~\ref{fig:overview} (b). 
    }%
    \Description{}
    \label{fig:transform-program}%
\end{figure}

Low-precision kernels in LLMs typically follow two steps before computation: (1) loading weights into on-chip memory (registers or shared memory) from global memory, and (2) casting and de-quantizing low-precision weights to high-precision (e.g., \code{float16}).
Efficient memory loading and casting are thus critical for performance.
\noindent\textbf{Efficient Low-Precision Weight Loading.} 
With the low-precision support as discussed in the previous subsection, we can use the \code{LoadGlobal} instruction to load low-precision tensors. 
However, directly loading in this way is inefficient due to multiple bitwise operations and non-coalesced memory accesses~\cite{cuda}. 
To address this, we transform the weight tensor layout in global memory to facilitate more efficient loading.
Without transformation, loading a register tensor with dtype \code{i6} and layout \code{local(2, 1).column\_spatial(4, 8).local(2, 1)} results in non-contiguous memory accesses, causing multiple \emph{memory access transactions}~\cite{cuda}. Moreover, extracting low-precision bits requires additional bitwise operations. To optimize this, we identify a \emph{compatible} tensor type with dtype \code{uint8} and layout \code{local(3).spatial(32)}, which retains the number of threads and thread local elements, while enabling efficient memory loading. As illustrated in Figure~\ref{fig:transform-program}, we partition the weight tensor \code{[K, N]} into tiles of shape \code{[BK, BN]}. Each tile is reinterpreted from \code{i6[BK, BN]} to \code{u8[BK * BN * 6 / 8]}~\lino{19} and stored contiguously~\lino{20}. 
This allows us to load tiles efficiently using the hardware-friendly instructions in Figure~\ref{fig:overview}~\lino{10, 11}, while also enabling pipelined asynchronous memory transfers like standard data types and avoiding any layout conversion that relies on shared memory.
This method generalizes to loading any low-precision tensor with arbitrary layout. 
More formally, given a tensor with $n$ bytes per thread and $T$ threads, we reinterpret it using dtype \code{uint8} and layout \code{local(n2).spatial(T).local(n1)}, where $n_1 = \gcd(n, 16)$ and $n_2 = n / \gcd(n_1, 16)$\footnote{$\gcd(a, b)$ represents the greatest common divisor of $a$ and $b$}. 

\noindent\textbf{Efficient Casting.} 
After loading, weights must be cast from low-precision to high-precision (e.g., \code{float16}) for computation, especially if hardware lacks native support for the given low-precision format. We leverage target-specific instructions for efficient vectorized casting. On CUDA, we use the \code{PRMT} (permute bytes in a 32-bit register), \code{LOP3} (arbitrary logical operation on three inputs), and bitwise instructions to execute casting with minimal overhead, as all operations are performed within registers and do not require any communication between threads.

\section{Implementation}
\system comprises five main components: a domain specific language (DSL) in Python, an intermediate representation (IR), optimization passes, a code generator, and a runtime system. The DSL enables developers to write \system programs in Python, which are then translated into the VM's IR for further processing.
Optimization passes refine the IR by eliminating redundancies and simplifying arithmetic expressions.
The code generator translates the optimized IR into Hidet IR~\cite{hidet}, a CUDA C-like intermediate representation.
Subsequently, we apply the transformations from Section~\ref{section:system:low-precision} to implement low-precision operations using standard precision types while preserving original semantics.
The final CUDA C code is generated from Hidet IR and compiled into a hardware binary using the \code{nvcc} compiler~\cite{cuda}. The runtime system manages dynamically loaded binaries and provides the execution environment.
The entire system consists of approximately 35K lines of Python and C++ code.

\subsection{Program Compilation and Runtime}
\label{section:appendix:implementation}
Tilus provides a user-friendly interface that allows programmers to write Tilus programs directly in Python, simplifying the integration of produced kernels with the rich deep learning ecosystem.
Given a program, we take several steps to compile it to GPU executable code. 

\noindent\textbf{Step 1: Global and Shared Memory Planning}
Each GPU kernel can use a shared memory space of a size determined at launch-time.
To simplify GPU programming, we allow the users to allocate shared memory multiple times in the program on demand. 
Thus, we need a shared memory planner to calculate the size of shared memory the Tilus program needs, and map the shared tensor to one region of the kernel's shared memory space. 
Similar to shared memory planning, we also require a global memory planner to manage the allocation of global memory shared by all thread blocks. 
We request Tilus's runtime system to allocate a workspace in global memory, enabling the kernel to use this workspace via the \code{AllocateGlobal} instruction during its execution.

\noindent\textbf{Step 2: Code Emitting for Each Instruction.}
We emit low-level GPU code for each Tilus instruction sequentially.
In our implementation, we use the Hidet IR~\cite{hidet} to represent the low-level GPU code. 
During this process, 
we perform instruction selection to choose the most efficient low-level instructions where feasible.
For example, we use \code{lds} PTX instruction~\cite{ptx} to load the data from shared memory to register. 
However, 
a more efficient PTX instruction \code{ldmatrix} could also be used if the layout of the loaded register tensor is compatible with the layout \code{spatial(8, 4).repeat(1, 4)}, which means it can be divided properly by the latter.
Besides, we also perform automatic vectorization for memory loading and storing instructions.
For example, we use vectorized instructions, such as \code{cp.async.v4}, \code{lds128}, and \code{ldg128}, to maximize memory access efficiency.

\noindent\textbf{Step 3: Lowering Low-Precision Data Types.}
After we emit thread-block-level instructions to the low-level IR, we will apply the passes that implement the the rules discussed in Section~\ref{section:system:low-precision-storage} to transform all low-precision operations in the low-level IR to corresponding operations on hardware-friendly types. 
In most cases, only the vectorized type casting from low-precision type to standard type (e.g., \code{float16}) will be applied since the memory loading of low-precision data will be replaced by standard types thanks to our layout formalization and register tensor reinterpretation.
Subsequently, we generate CUDA code (for NVIDIA GPUs) from the low-level IR and then use the \code{nvcc} compiler to produce the hardware binary that can be dynamically loaded.

\noindent\textbf{Step 4: Loading by Runtime System.} 
The compiled binary is dynamically loaded by the Tilus's runtime system.
The runtime system also maintains internal states to serve the kernel execution: 1) a workspace memory allocated on-demand, which compiled kernels can request via \code{AllocateGlobal} instruction; 
2) an execution context that stores the CUDA stream on which the kernel are launched; and 3) the kernels cached in memory to avoid unnecessary recompilation.

\begin{figure*}[th]%
    \centering
    \includegraphics[width=1.0\linewidth]{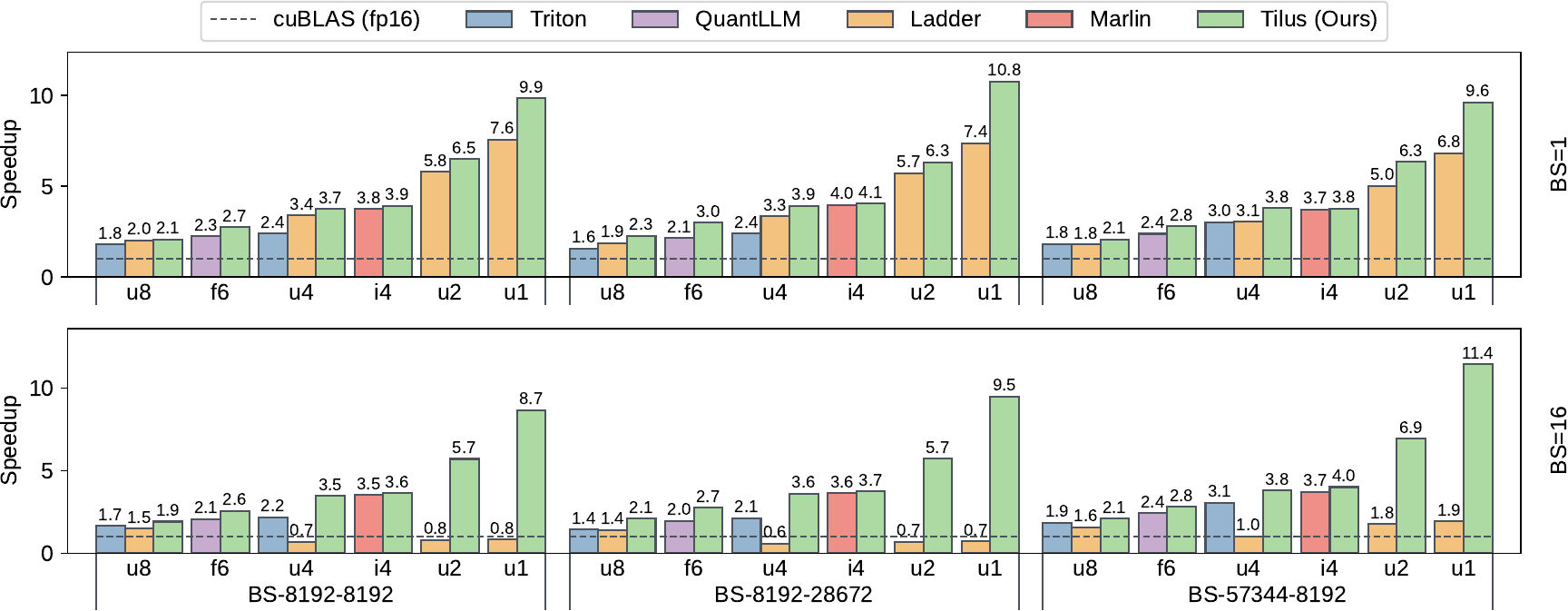} 
    \caption{
    Speedup of low-precision kernels in Triton, QuantLLM, Ladder, and \system (Ours) compared against the standard half-precision kernel from cuBLAS. Benchmarked data types include \texttt{uint8} (u8), \texttt{f6e3m2} (f6), \texttt{int4} (i4), \texttt{uint4} (u4), \texttt{uint2} (u2), and \texttt{uint1} (u1). Each workload (BS-N-K) corresponds to a matrix multiplication in Llama-3.3-70B, with batch sizes 1 and 16.
    }%
    \Description{}
    \label{fig:exp_operators}%
\end{figure*}

\section{Evaluation}

\subsection{Experimental Setup}

\noindent\textbf{Workloads.}
We benchmark three representative LLMs with varying model sizes: Gemma-2-9B~\cite{gemma}, QWen2.5-32B~\cite{qwen2}, and Llama-3.3-70B-Instruct~\cite{llama3}.
Both the prefill and decode stages are evaluated. For operator-level analysis, we focus on matrix multiplication kernels extracted from these models. 
\system supports all kernels supported by Triton in principle, but we focus on quantized matmul in this work.

\noindent\textbf{Baselines.}
We compare our approach, \vm, against the vendor library cuBLAS~\cite{cublas}, state-of-the-art DL compilers Triton~\cite{triton} and Ladder~\cite{ladder}, and hand-crafted kernels QuantLLM \cite{quant-llm} and Marlin~\cite{marlin}.
Auto-tuning for Triton~\cite{triton} and Ladder~\cite{ladder} was enabled, while QuantLLM~\cite{quant-llm} used its heuristic policy to select kernel hyperparameters.
For end-to-end evaluations, we integrate our quantized kernels into the state-of-the-art LLM serving framework vLLM~\cite{vllm} and compare them against vLLM~\cite{vllm} and Ladder~\cite{ladder} in end to end execution.
The specific versions of the tools are: vLLM \code{v0.5.3}, Triton \code{v3.1.0}, bitblas \code{v0.0.1.dev15} (Ladder), QuantLLM with commit \code{9802c5a}, and Marlin \code{v0.1.1}.

\noindent\textbf{Hardware Configuration.}
Experiments were primarily conducted on a server equipped with an NVIDIA L40S GPU (48 GiB), with GPU driver \code{565.57.01} and CUDA Toolkit \code{12.6.3}.
Benchmarks were also performed on NVIDIA A100 and H100 GPUs to demonstrate the general applicability of our approach across different hardware platforms.

\noindent\textbf{Experimental Protocol.}
For operator experiments, each kernel was executed $50$ times, while for model experiments, each model was executed $10$ times. 
In both cases, latency was measured using CUDA Events~\cite{cuda}, and the median latency was reported. To eliminate artifacts from consecutive runs, the L2 cache was cleared before each execution. 

\subsection{Performance of Low-Precision Kernels}
A single virtual machine program template is implemented to support matrix multiplication with all quantized types, taking tile sizes as tunable hyperparameters. We denote the performance of this auto-tuned program as \system in the evaluation.
Figure~\ref{fig:exp_operators} compares the speedup of Triton~\cite{triton}, Ladder~\cite{ladder}, QuantLLM~\cite{quant-llm}, Marlin~\cite{marlin}, and \system (ours) against cuBLAS~\cite{cublas} for various low-precision matrix multiplications: \dtype{uint8} (u8), \dtype{float6\_e3m2} (f6), \dtype{uint4} (u4), \dtype{int4} (i4), \dtype{uint2} (u2), and \dtype{uint1} (u1). While each baseline supports a limited set of quantized data types, \system consistently achieves speedups across all cases.
For small batch sizes, the primary bottleneck is loading weights from global memory to registers for computation on SIMT or Tensor Cores. Triton struggles here due to costly layout conversions after weights are loaded into registers. Although changing layout in global memory could mitigate this, Triton's programming model lacks explicit layout control, making such optimizations infeasible.
Ladder improves upon Triton by modifying data layouts in global memory, avoiding redundant conversions. However, it lacks critical optimizations such as software pipelining~\cite{cutlass, alcop}, and its type-level packing limits efficient support for arbitrary bit widths, leading to underutilized memory bandwidth.
Expert-crafted kernels from QuantLLM~\cite{quant-llm} and Marlin~\cite{marlin} are optimized for specific quantization schemes but lack flexibility and maintainability. 
In contrast, \system consistently outperforms all baselines using a single parameterized Tilus program template, which efficiently supports a full range of quantization types through a well-abstracted programming model.

\subsection{Arbitrary Data Type Support}
\begin{figure}[th]%
    \centering
    \includegraphics[width=0.98\linewidth]{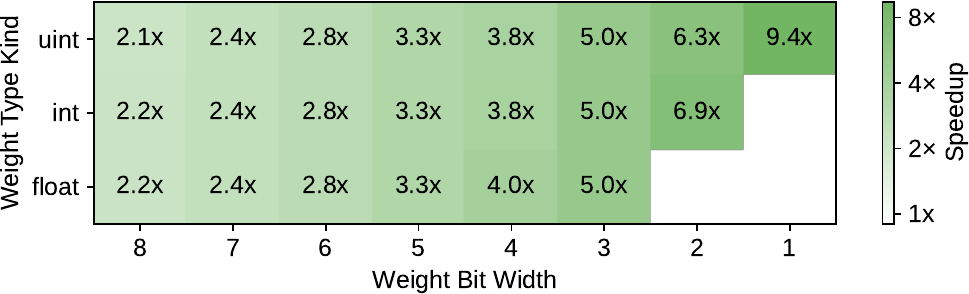} 
    \caption{
    Speedup of quantized matrix multiplication compared against the cuBLAS FP16 kernel. A full spectrum of quantized data types is evaluated.
    }%
    \Description{}
    \label{fig:exp_coverage}%
\end{figure}

Tilus supports low-precision matrix multiplications of the form \texttt{matmul(A, B)}, where operand A can have data types with 32, 16, or 8 bits, and weight B supports a wide range of bit widths, from 32 bits down to 1 bit. Standard data types such as \texttt{float32}, \texttt{float16}, and \texttt{int8} are supported, along with customized low-precision types with fewer than 8 bits, which include signed integers, unsigned integers, and floating-point formats with arbitrary exponent and mantissa distributions.
Leveraging the algebraic layout system (Section~\ref{section:system:layout} and~\ref{section:low-precision:efficient-support-in-llm}), Tilus enables efficient memory access for low-precision data. 
Figure~\ref{fig:exp_coverage} illustrates the speedup achieved for the full spectrum of quantized weight data types: \code{uint1} to \code{uint8}, \code{int2} to \code{int8}, and \code{float3} to \code{float8}. 
Representative exponent-mantissa distribution of floating-point data types such as \texttt{e4m3}, \texttt{e3m3}, \texttt{e3m2}, \texttt{e2m2}, \texttt{e2m1}, and \texttt{e1m1} are chosen. 
Each row represents the type kind (e.g., unsigned integer, signed integer or floating data type) while each column represents the bit width. 
Using matrix multiplication dimensions of BS=16, K=8192, and N=57344 the results demonstrate substantial speedups.
These findings validate Tilus's effectiveness in supporting arbitrary low-precision types with high efficiency, making it a robust solution for low-precision computations in modern GPUs. 
Notably, all kernels are generated from the same program template by parameterizing tile sizes, which limits the required programming effort.
There are around 200 configurations per operator, and it takes around one minute to compile.
We used \code{float16} as the activation data type in the experiment and we also support \code{bfloat16} and \code{int8}.

\subsection{End-to-End Performance}
\begin{figure}[th]%
    \centering
    \includegraphics[width=1.0\linewidth]{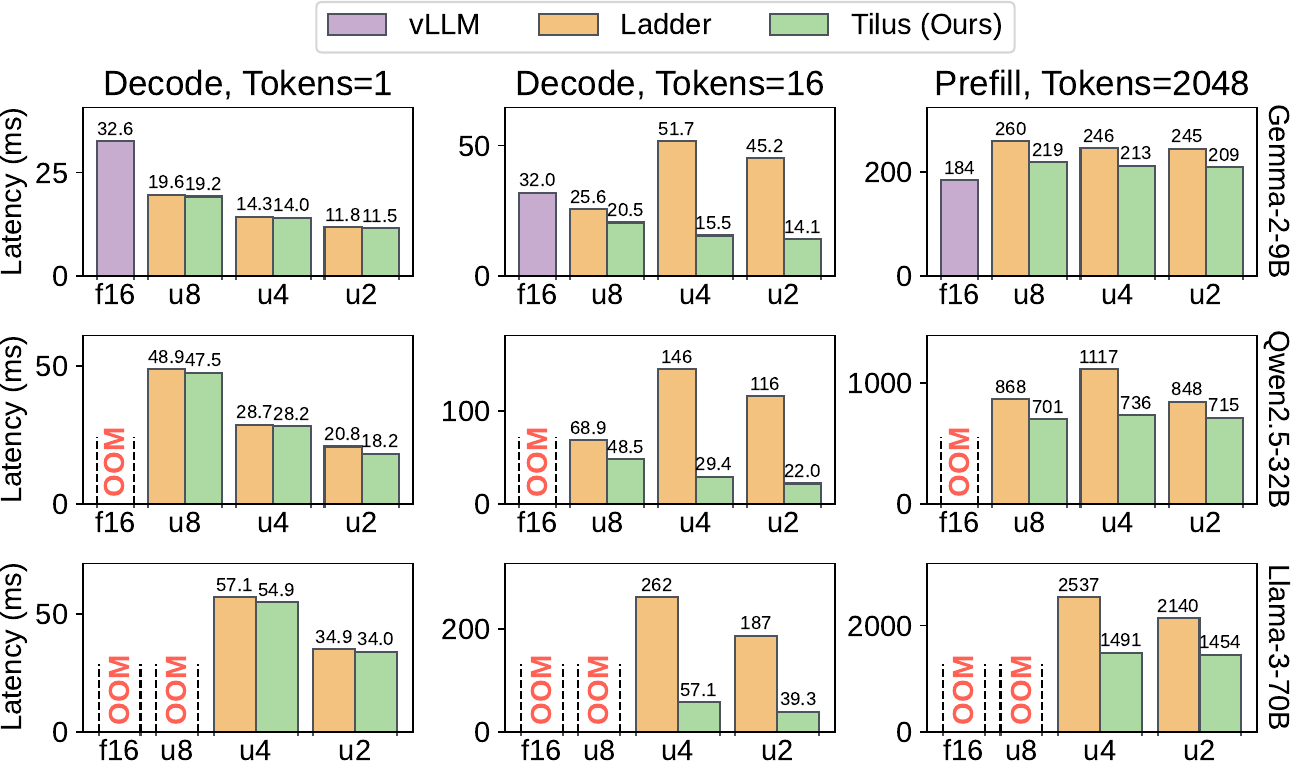} 
    \caption{
    End-to-end performance across representative LLMs. The first two columns correspond to the decode stage with 1 and 16 tokens, respectively, while the third column shows latency for the prefill stage with 2048 prompt tokens.
    }%
    \Description{}
    \label{fig:exp_models}%
\end{figure}

We evaluated the end-to-end performance of representative LLMs: Gemma-2-9B~\cite{gemma}, QWen-2.5-32B~\cite{qwen2}, and Llama-3.3-70B~\cite{llama3}, across both prefill and decode stages. 
The prefill stage processes all prompt tokens at once, generating the kv-cache for subsequent token generation. The decode stage then iteratively generates one token at a time.
Prefill latency determines the time-to-first-token (TTFT), while decode latency impacts the speed of subsequent token generation. Both stages are critical for optimizing user experience and system utilization. 
Contiguous batching~\cite{orca, vllm} was used to efficiently batch multiple decode requests.
Figure~\ref{fig:exp_models} shows the latency of both stages across these models. Our method consistently outperforms Ladder~\cite{ladder}, particularly in the decode stage 
for batch sizes greater than one
(middle column of Figure~\ref{fig:exp_models}). Analysis of Ladder's generated kernels revealed suboptimal use of CUDA Cores for 1–15 tokens and Tensor Cores for 16 or more tokens, as key optimizations like software pipelining~\cite{alcop} and k-dimension parallelization~\cite{stream-k} were not implemented, leading to poor performance.
For the prefill stage, quantized weights are decoded to float16, and computations are performed using standard f16xf16 matrix multiplication kernels,
as computation becomes the bottleneck at this stage. Our efficient handling of quantized weight layouts ensures minimal overhead for decoding, contributing to the superior performance observed.

\subsection{Case Studies}
\subsubsection{Speedup over Different Hardware}
\begin{figure}[th]%
    \centering
    \includegraphics[width=1.0\linewidth]{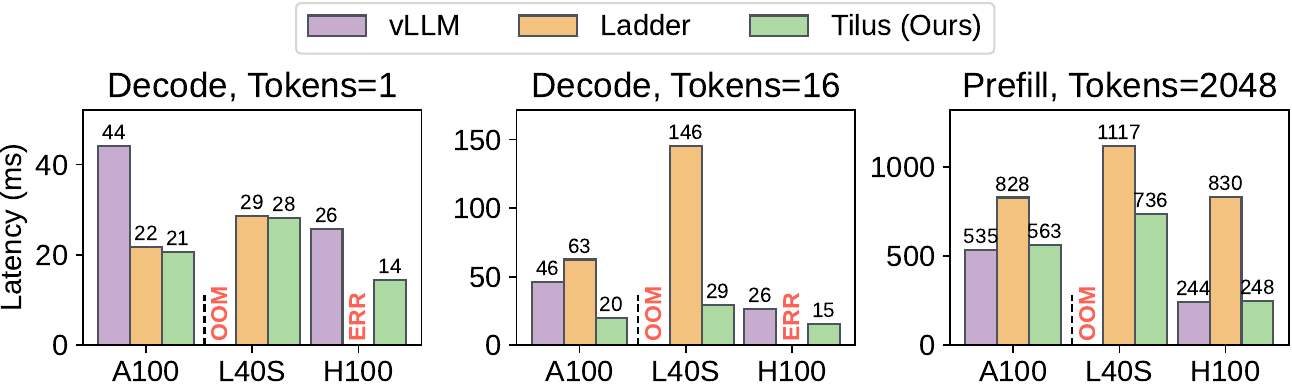} 
    \caption{
    End-to-end performance of the QWen2.5-30B model across NVIDIA A100, L40S, and H100 GPUs. The weight data types for vLLM, Ladder, and \system are \texttt{float16}, \texttt{uint4}, and \texttt{uint4}, respectively. OOM indicates out-of-memory error, and ERR indicates a runtime error.
    }
    \Description{}
    \label{fig:exp_hardware}%
\end{figure}
We evaluate the end-to-end performance of the QWen2.5-30B model on NVIDIA A100, L40S, and H100 GPUs, which correspond to the Ampere, Ada Lovelace, and Hopper architectures, respectively.
Figure~\ref{fig:exp_hardware} presents a performance comparison of vLLM~\cite{vllm} (float16), Ladder~\cite{ladder} (uint4), and \system (uint4, ours) across the decode and prefill stages.
On the Hopper architecture (H100), Ladder is unable to generate valid kernels, leading to a CUDA error ('an illegal instruction was encountered'), which we denote as ERR in the figure.
On the L40S GPU, vLLM~\cite{vllm} exceeds the available 48 GiB DRAM capacity, leading to out-of-memory (OOM) errors.
In all other configurations, \system consistently outperforms Ladder across all GPUs and both processing stages, highlighting its robust performance and adaptability across architectures.

\subsubsection{Speedup over Different Batch Sizes}
\begin{figure}[th]%
    \centering
    \includegraphics[width=1.00\linewidth]{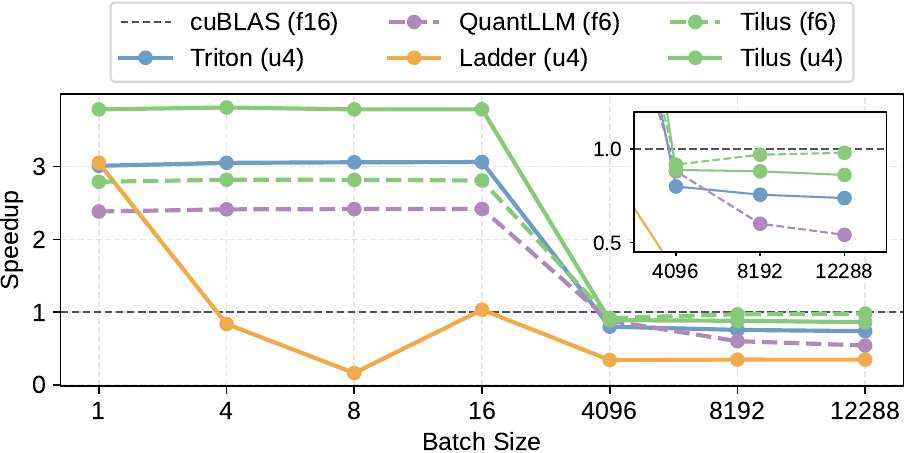} 
    \caption{Speedup of quantized matmuls across different batch sizes from both prefill and decode stages.}%
    \Description{}
    \label{fig:exp_batch_sizes}%
\end{figure}
We analyze the relationship between speedup and batch size by benchmarking matrix multiplication performance under different batch sizes. For the decode stage, we evaluate batch sizes of 1, 4, 8, and 16, while for the prefill stage, we use batch sizes of 4096, 8192, and 12,288. The batch size corresponds to the number of tokens processed in one step. In the prefill stage, it equals the sum of sequence length of all requests, while in decode stage, it equals to the number of requests since each request only generates one token each time. Experiments are conducted on Llama-3.3-70B-Instruct~\cite{llama3} model with quantized data types \code{float6\_e3m2} (f6) and \code{uint4} (u4), using $k=8192$ and $n=57344$.
As shown in Figure~\ref{fig:exp_batch_sizes}, \vm consistently outperforms baselines across all batch sizes that are used in both decode and prefill stages of LLM serving.

\section{Related Work}

Many deep learning compilers adopt loop-oriented scheduling~\cite{tvm,halide} and build auto-tuning frameworks on top of it~\cite{halide_autoscheduler,autotvm,tensor_comprehension,chameleon,bolt,dietcode,ansor,amos,flextensor,meta-schedule,tensorir,ladder}. In contrast, \system employs a procedure-oriented approach that better models GPU hardware, improving programmability and flexibility.
Beyond loop-oriented scheduling, tensor programs are often optimized using vendor libraries (e.g., cuBLAS~\cite{cublas}), predefined templates for efficient matrix multiplication~\cite{cutlass}, hardware-aware tiling strategies~\cite{roller}, and domain-specific compilers for linear algebra~\cite{tensorflow-xla}. While these methods prioritize performance, they lack extensibility for arbitrary low-precision data types.
Other research focuses on optimizing irregular or ragged tensor programs~\cite{freetensor,CoRa}, operator fusion~\cite{astitch,apollo}, dynamic shape handling~\cite{nimble_haichen,disc,cortex,dietcode}, and scheduling independent operators~\cite{rammer,ios,taso}. 
Prior works~\cite{abq-llm, arb-prec, APNN-TC, any-precision} endeavor to support quantization with arbitrary bit-width by independently processing different bits. However, these methods are primarily limited to integer quantization and often exhibit suboptimal execution performance. PartIR~\cite{part-ir} also introduces a layout system for tiles; however, its abstraction level is higher than Tilus's, and it does not detail the assignment of tile elements to threads. ExTensor~\cite{ExTensor} and FuseMax~\cite{FuseMax} leverage einsum to define computations, implicitly determining memory access patterns. Nevertheless, they do not specify how tile elements are distributed among GPU threads. Tilus can also accommodate codebook quantization, such as LCQ~\cite{lcq}, to facilitate low-precision computation through the addition of a new lookup instruction.
Microscaling data types~\cite{mx-format} can be thought as a more fine-grained quantization thus we could also support it.
These techniques are complementary to our focus on efficient low-precision computation.
Triton~\cite{triton} introduces a tile-based programming model. However, it lacks explicit support for low-precision data types and does not expose the GPU memory hierarchy, limiting optimization opportunities. Similarly, Hidet~\cite{hidet}, which serves as our backend, does not provide built-in support for low-precision types.
Graphene~\cite{graphene} presents an intermediate representation (IR) with a layout representation. Unlike Graphene’s focus on strides and computation, our algebraic layout system emphasizes hierarchical organization. In fact, we can express Graphene’s layout representation as a primitive component within our system.

\section{Conclusion}

We introduced \system, a tile-level GPGPU programming language designed to expose shared memory and registers to developers, which enables the creation of efficient low-precision kernels for LLM serving. \system features an algebraic layout system for managing tensor distribution across threads, a thread-block-level programming model with fine-grained memory management, and comprehensive support for sub-byte data types, enabling arbitrary precision from 1 to 8 bits.
Our experimental results demonstrate substantial performance gains over state-of-the-art approaches, showcasing the flexibility and scalability of our method. This work establishes a foundation for efficient and extensible LLM inference, paving the way for future optimizations in emerging hardware, advanced quantization techniques, and diverse low-precision formats.

\section*{Acknowledgement}

We would like to thank the members of the EcoSystem research laboratory at the University of Toronto for their feedback on the early manuscript. 
We also thank the anonymous ASPLOS reviewers and our shepherd, Jacques Pienaar, for their valuable feedback and suggestions, as well as the artifact evaluation reviewers for reproducing our experiments. The authors affiliated with the University of Toronto were supported by the Canada Foundation for Innovation JELF grant, the NSERC Discovery grant, the Google Scholar Research Award, the VMware Early Career Faculty Grant, and the Ontario Early Researcher Award.

\appendix

\section*{Appendix}

\section{Artifact Appendix}

\subsection{Abstract}

We provide artifacts to reproduce all experimental results discussed in the evaluation section. These artifacts include the compiler implementation, kernels in our domain-specific languages, and scripts for running experiments and plotting figures. To simplify reproduction, all necessary dependencies are packaged within a Docker image. A Dockerfile is also included to demonstrate the image building process. We do not provide the models directly; instead, they and their metadata will be automatically downloaded from Hugging Face Hub when the experiment scripts are launched.

\subsection{Artifact check-list (meta-information)}

{\small
\begin{itemize}
  \item {\bf Model: Gemma, Llama, Qwen}
  \item {\bf Run-time environment: Linux, CUDA}
  \item {\bf Hardware: NVIDIA L40s}
  \item {\bf Metrics: Latency (ms), Speedup}
  \item {\bf Output: Efficient kernels, Numerical results}
  \item {\bf Experiments: Automated scripts in docker}
  \item {\bf How much disk space required (approximately)?: 25 GiB}
  \item {\bf How much time is needed to prepare workflow (approximately)?: 10 minutes}
  \item {\bf How much time is needed to complete experiments (approximately)?: 3 hours}
  \item {\bf Publicly available?: Yes}
  \item {\bf Code licenses (if publicly available)?: Apache 2.0}
  \item {\bf Workflow automation framework used?: Docker}
  \item {\bf Archived (provide DOI)?:\\ \url{https://doi.org/10.5281/zenodo.16756859}}
\end{itemize}
}

\subsection{Description}

\subsubsection{How to access}

We have open-sourced our artifacts at \url{https://github.com/yaoyaoding/tilus-artifacts}.
To pull the Docker image and perform experiments, follow the guide in the \code{README.md} file. The code itself is several megabytes, while the Docker image is approximately 21 GiB. We only fetch model meta-information (e.g., number of layers, layer size) from Hugging Face Hub, and dummy weights are used; therefore, the models do not consume significant disk space.

\subsubsection{Hardware dependencies}

Our experiments were primarily conducted on NVIDIA L40s. To perform a hardware ablation study, we also ran some experiments on NVIDIA A100 and NVIDIA H100. Any NVIDIA GPU with compute capability $\ge 8.0$ should be able to run our artifacts and observe speedup, though the specific numbers might vary slightly.

\subsubsection{Software dependencies}

We provide a Docker image with all software dependencies pre-installed. Therefore, only the following software is required:
\begin{itemize}
    \item NVIDIA GPU driver $\ge$ \code{565.57.01}
    \item NVIDIA container toolkit
    \item Docker 
\end{itemize}

We have the following packages pre-installed:
\begin{itemize}
    \item PyTorch \code{v2.5.1}
    \item Triton \code{v3.1.0}
    \item BitBLAS \code{v0.0.1.dev15}
    \item Marlin \code{v0.1.1}
    \item vLLM \code{0.7.3}
\end{itemize}

\subsubsection{Data sets}

We use dummy inputs and weights because we are solely focused on system performance, which is independent of the input and weight content.

\subsubsection{Models}

The artifact uses three models for end-to-end evaluation: Gemma-2-9B, QWen-2.5-32B, and Llama-3.3-70B. The meta-information of these models will be automatically fetched from Hugging Face Hub (some may require a Hugging Face token).

\subsection{Installation}

\noindent First, clone the artifact Git repository:

\begin{flushleft}
{\scriptsize\texttt{git clone https://github.com/yaoyaoding/tilus-artifacts.git tilus}}
\end{flushleft}

\noindent Then, install Docker and the NVIDIA Container Toolkit by following the \code{README.md} in the artifact.

\subsection{Experiment workflow}

All experiments can be executed with:
\begin{flushleft}
{\scriptsize\texttt{bash run.sh}}
\end{flushleft}

\noindent This command will create a Docker container and sequentially run all experiments within it.

\subsection{Evaluation and expected results}

Upon completion of the experiments, a folder named \code{results} or \code{precompiled-results} will be created under the artifact directory. This folder will contain four figures of the evaluation results, corresponding to those presented in the evaluation section.

\subsection{Methodology}

Submission, reviewing and badging methodology:

\begin{itemize}
  \item \href{https://www.acm.org/publications/policies/artifact-review-and-badging-current}{https://www.acm.org/publications/policies/artifact\\-review-and-badging-current}
  \item \url{https://cTuning.org/ae}
\end{itemize}

\bibliographystyle{ACM-Reference-Format}
\balance
\bibliography{
    references/llm, 
    references/quantization,
    references/system
}

\end{document}